\documentclass[lettersize,journal]{IEEEtran}
\usepackage{amsmath,amsfonts}
\usepackage{algorithmic}
\usepackage{algorithm}
\usepackage{array}
\usepackage[caption=false,font=normalsize,labelfont=sf,textfont=sf]{subfig}
\usepackage{textcomp}
\usepackage{stfloats}
\usepackage{url}
\usepackage{verbatim}
\usepackage{graphicx}
\usepackage{cite}
\usepackage{amsmath}
\usepackage{booktabs}
\usepackage{amssymb}
\usepackage{multirow}
\usepackage{bbding}
\usepackage{xspace} 
\usepackage{xcolor}
\usepackage[normalem]{ulem} 

\makeatletter
\DeclareRobustCommand\onedot{\futurelet\@let@token\@onedot}
\def\@onedot{\ifx\@let@token.\else.\null\fi\xspace}

\def\eg{e.g\onedot} 

\def\ie{i.e\onedot}

\def\etc{etc\onedot}

\makeatother

\def\method{CFSynthesis} 

\newcommand{\zp}[1]{#1} %
\newcommand{\ly}[1]{#1} %
\newcommand{\zpn}[1]{#1} %

\begin{document}

\title{CFSynthesis: Controllable and Free-view 3D Human Video Synthesis}


\author{Liyuan Cui, Xiaogang Xui\textsuperscript{‡}, Wenqi Dong, Zesong Yang, Hujun Bao, \textit{Member, IEEE}, and Zhaopeng Cui\textsuperscript{$\dagger$}, \textit{Member, IEEE}
    \IEEEcompsocitemizethanks{
    \IEEEcompsocthanksitem ‡ Project lead.
    \IEEEcompsocthanksitem $\dagger$ Corresponding author.
    \IEEEcompsocthanksitem Liyuan Cui, Wenqi Dong, Zesong Yang, Hujun Bao, and Zhaopeng Cui are with the State Key lab of CAD\&CG, College of Computer Science, Zhejiang University. Email: \{cuiliyuan, dongwenqi, zesongyang0\}@zju.edu.cn, bao@cad.zju.edu.cn, zhpcui@zju.edu.cn.
    \IEEEcompsocthanksitem Xiaogang Xu is with the department of computer science and engineering, the Chinese University of Hong Kong, Hong Kong, China. Email:xiaogangxu00@gmail.com.
    }
}


\markboth{Journal of \LaTeX\ Class Files,~Vol.~14, No.~8, August~2024}%
{Shell \MakeLowercase{\textit{et al.}}: A Sample Article Using IEEEtran.cls for IEEE Journals}


\maketitle

\begin{figure*}[htbp] 
    \centering
    \includegraphics[width=\textwidth]{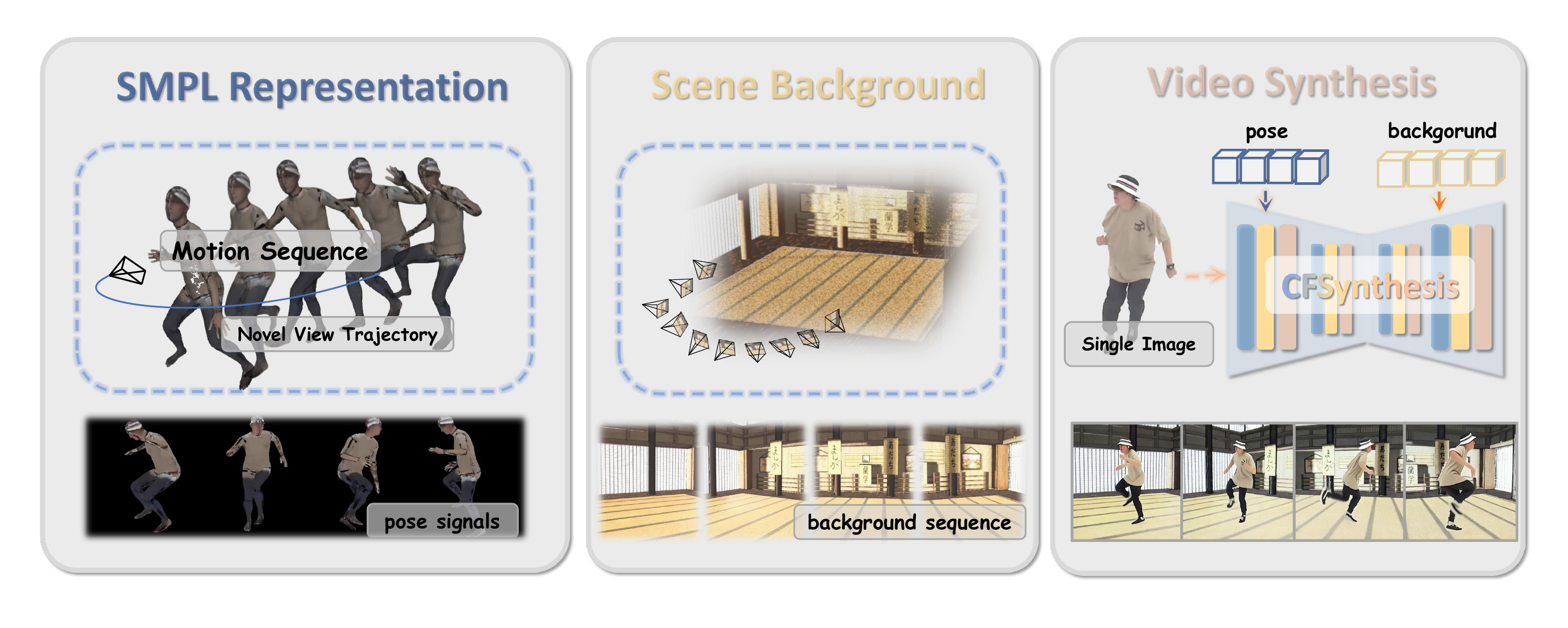}
     \vspace{-0.45in}
    \caption{
     \textbf{CFSynthesis}. Given a single reference image, \method\ can synthesize human videos driven by a texture-based SMPL representation derived from 3D pose estimation or generation. It also integrates user-desired scenes as controllable attributes, enabling the generation of lifelike 3D motion videos with \zp{varying} backgrounds in free-view.
    }
    \label{fig:teaser}
\end{figure*}

\begin{abstract}
Human video synthesis aims to create lifelike characters in various environments, with wide applications in VR, storytelling, and content creation. While 2D diffusion-based methods have made significant progress, 
\zpn{they struggle to generalize to complex 3D poses and varying scene backgrounds.}
\zpn{To address these limitations,} we introduce \method, a novel framework for generating high-quality human videos with customizable attributes, including identity, motion, and scene configurations. Our method leverages a texture-SMPL-based representation to ensure consistent and stable character appearances across free viewpoints. Additionally, we introduce a \zpn{novel} foreground-background separation strategy that effectively \zpn{decomposes the scene }
as foreground and background, 
\zpn{enabling seamless integration of user-defined backgrounds.}
Experimental results on multiple datasets show that \method\ not only achieves state-of-the-art performance in \zpn{complex human} 
animations but also adapts effectively to 3D motions in free-view and user-specified scenarios. 
\end{abstract}

\begin{IEEEkeywords}
Human Animation, Stable Diffusion, Video Control.
\end{IEEEkeywords}

\section{Introduction}
\IEEEPARstart{F}{ree}-view human video synthesis using controllable signals~(\eg, the motion, identity, \etc) is a complex yet significant generative task in computer vision and computer graphics. It has immense potential for applications in various fields, including virtual reality experiences~\cite{maiorca2024selfavatar, patel2024fast, yan2023gazemodiff}, interactive narratives~\cite{windle2024llanimation, li2024sparse}, and digital content creation~\cite{zheng2024laser,xi2024maskel}.

\zp{Earlier works mostly focus on human animation and normally}
use data-driven approaches with Generative Adversarial Networks (GANs)~\cite{gan, tian2021good, wang2021oneshot}. Due to the limited generative capabilities of GANs, these methods typically involve warping the source image to \zp{align with} the target signal in the latent expression space. Warping can be achieved through \zp{both} implicit and explicit motion modeling, such as using 2D optical flow~\cite{ren2021pirenderer, zhao2022thinplate} or 3D deformation fields~\cite{mallya2022implicit, wang2021oneshot}. However, these approaches struggle with interpolating some occluded parts, as the corresponding warping errors cannot be fully eliminated. These errors often result in \zp{visual} artifacts, \zp{such as distortions in the characters' identities, which heavily degrade the quality of the synthesized videos\cite{wang2018vid2vid, tulyakov2018mocogan}.}

Recently, with the rapid development of diffusion models and ControlNet\cite{zhang2023adding}, 
\zp{significant progress has been made in leveraging motion signals such as depth maps~\cite{dreamoving}, skeletons~\cite{disco}, and dense motion flows~\cite{dreampose}.}
While achieving general pose control, these methods exhibit deficiencies in preserving human appearance details and fidelity.
Therefore, \zp{latest} approaches such as \cite{magicanimate, anymateanyone, chang2023magicpose} \ly{have utilized a replica of U-Net, to encode the reference image}
in a consistent feature space through spatial attention, thereby enhancing the preservation of appearance details.  
\zp{Despite} achieving higher video quality, these methods \zp{still} primarily focus on character animation within basic 2D motions, characterized by \ly{restricted} poses and fixed backgrounds.
\zpn{This further} limits their ability to model complex human movements in 3D space and 
\ly{insert a brand new background environment.}
\ly{The reliance on inadequate 2D expressions fails to faithfully maintain the stability of appearance amid variations in viewpoints and scenes.}

\begin{table}[!t]
    \centering
    \caption{\textbf{Function comparison between \method\ and recent human video generation methods}. Here, ``animation" refers to foreground motion, ``free-view" represents novel perspectives distinct from the input view, and ``background" refers to the insertion of user-specified dynamic backgrounds.}
    \label{tab:setting}
    \begin{tabular}{p{2.7cm} >{\centering\arraybackslash}p{1cm} >{\centering\arraybackslash}p{1.6cm} >{\centering\arraybackslash}p{1.4cm}} 
        \toprule
        Methods & Animation & Free-View & Background \\
        \midrule
        DreamPose\cite{dreampose}  & \Checkmark    & \XSolidBrush & \XSolidBrush \\
        MagicAnimate\cite{magicanimate}   & \Checkmark   & \XSolidBrush & \XSolidBrush \\
        AnimateAnyone\cite{anymateanyone} & \Checkmark  & \XSolidBrush & \XSolidBrush \\
        Champ\cite{champ}   & \Checkmark   & \XSolidBrush & \XSolidBrush \\
        Human4Dit\cite{shao2024human4dit360degreehumanvideo} & \Checkmark  & \Checkmark   & \XSolidBrush \\
        Ours   & \Checkmark    & \Checkmark   & \Checkmark   \\
        \bottomrule
    \end{tabular}
    
\end{table}

To facilitate lifelike and flexible user-controlled videos in demanding scenarios, such as extreme 3D motions and customizable backgrounds, we need a unified framework that offers flexibility 
in human animation, versatility in managing free-view 3D motions, and adaptability to interactive real-world environments. However, 
\zp{achieving this goal presents two key challenges:}
 
\begin{itemize}
    \item The framework should \zp{be able to} transfer the human appearance from a single reference image to diverse human poses and novel views.    
    \item The framework should effectively decouple the spatial relationships between the character and the scene background, 
    \zp{enabling video synthesis with entirely novel scene backgrounds. }
\end{itemize}

\zp{In this paper, we present \method, a novel system for controllable and free-view 3D human video generation,} which has two key designs: a texture-SMPL-based pose representation that ensures view consistency across 360-degree projections and a foreground-background separation learning strategy that utilizes the background as control signals for the synthesis process. 
\zp{As shown in Table~\ref{tab:setting}, these innovations enable \method\ to go beyond traditional human animation, extending its capabilities to more advanced human video synthesis tasks, such as free-view motion transfer and user-desired scene insertion. }

\zp{To establish a consistent multi-view synthesis, constructing or encoding a complete human body information plays a key role.}
Existing 3D methods \cite{hu2024gaussianavatar, li2024animatable, liu2021neural, peng2021animatable, weng2022humannerf} often require capturing multiple views for each training case, significantly limiting their capacity to model diverse human representations efficiently. 
\ly{2D techniques\cite{magicanimate, anymateanyone, chang2023magicpose} based on pretrained Stable Diffusion (SD) have overcome efficiency limitations,}
but they attempt to generate novel views \ly{solely} using inadequate 2D pose signals and abstract appearance features extracted by U-Net across large datasets, \ly{rather than employing geometric methods to learn the transition from input images to free-view perspectives.}
\zp{Based on all these observations,} we propose a texture-SMPL-based pose representation that provides intuitive texture priors. \ly{We inherit the
network design from SD and integrate such 3D priors to SD to ensure}
pixel uniformity across perspectives. 
The SMPL is a statistically accurate 3D human body representation. When combined with pixel-level priors, it can effectively overcome perspective limitations through camera projection, and guide the abstract appearance characteristics to fill in the current novel view, ensuring multi-view consistency 
\zp{without relying on extensive well-captured video sequences for training}. 
To improve the extraction of \ly{this structured pose information}, we designed a pose extractor that injects pose signals into the denoising processing.

\zp{Furthermore, }in contrast to previous studies \cite{3DControl, shao2024human4dit360degreehumanvideo, anymateanyone, magicanimate} 
that attempt to learn the complete frame features without decomposing essential attributes like video backgrounds, we propose \zp{an approach} that explicitly separates these components for improved representation.
\zp{Specifically, we propose to decompose}  the frame into different spatial components: foreground and background. This decoupling enables richer contextual information and serves as effective control signals for the synthesis process, allowing for more flexible and comprehensive user control. \zp{To achieve this,} we develop a foreground encoder to inject precise appearance information into the latent diffusion model at various resolutions, complemented by a learnable background encoder to accurately obtain scene embeddings during the SD decoding process. Additionally, we explored a robust fusion \zp{technique} to mitigate foreground edge flickering issues commonly encountered in prior works\cite{disco}.

We conduct extensive experiments on widely used 2D dance datasets~\cite{TikTok}, 3D motion datasets~\cite{aist-dance-db}, and in-the-wild 4D data. The experimental results demonstrate that our framework achieves state-of-the-art performance across these diverse datasets.
Our contributions can be summarized as follows:
\begin{itemize}
    \item We propose a novel framework, \method, which achieves high-quality \zp{human video synthesis} 
    while offering flexible user controls to enable the synthesis of complex motions, free-viewpoint transfer, and \zp{insertion with new scene backgrounds.} 
    \zp{\item We introduce an effective 3D expression with texture priors to maintain multi-view consistency to express complex motions across varying viewpoints without relying on extensive videos for training.}
    \zp{\item We accurately model the spatial relationships and propose a foreground-background separation learning strategy, which allows users to control both characters and scenarios.}
    \zp{\item Extensive experiments on multiple datasets demonstrate the effectiveness and superiority of our framework in comparison to existing methods.}
\end{itemize}

\begin{figure*}[htbp] 
    \includegraphics[width=\linewidth]{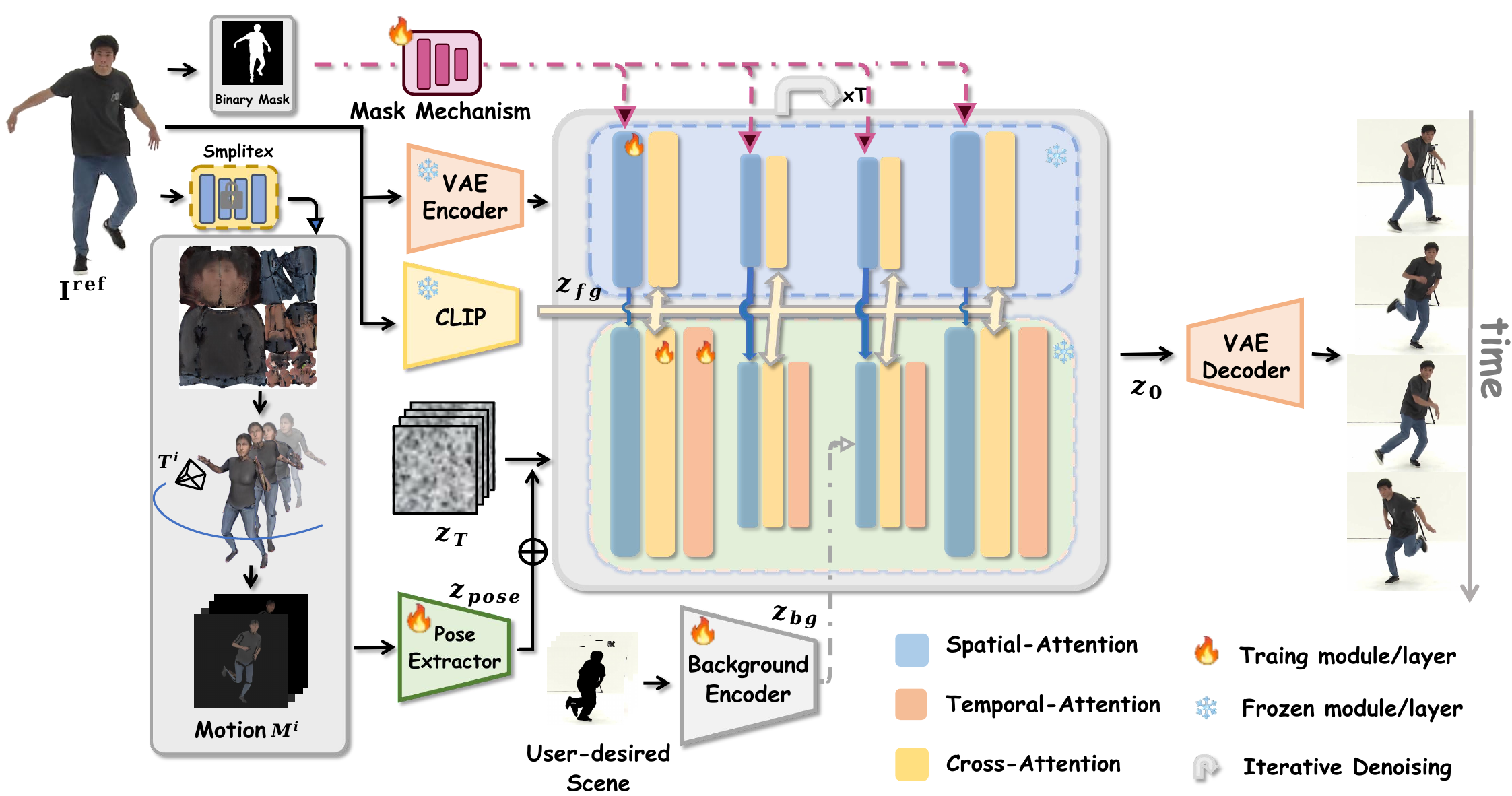}
    \caption{
    \textbf{An overview of the proposed framework.} 
     \ly{\method\ first warps an estimated texture map on the given 3D motion sequence and projects it to 2D space through camera pose ${T}^{i}$ to get the SMPL representation ${M}^{i}$}. It is then encoded as pose signals $\boldsymbol{z}_{pose}$. The foreground and background are separately encoded as $\boldsymbol{z}_{fg}$ and $\boldsymbol{z}_{bg}$, respectively, and are recomposed during the decoder stage using a masking mechanism. These components collaboratively guide the original latent code $\boldsymbol{z}_{0}$ for the target frame. In the U-Net architecture, the training/frozen strategy is uniform across all layers, and here we only illustrate the first layer.
    }
    \label{fig:pipeline}
\end{figure*}   

\section{Related Work}
\noindent\textbf{Stable Diffusion and ControlNet.}
In the field of text-to-image generation, diffusion-based methods~\cite{balaji2023ediffi,nichol2022glide, saharia2022photorealistic} have achieved significant success, leading to a proliferation of related works. Latent diffusion models~\cite{rombach2021highresolution} proposed a denoising method in the latent space, effectively reducing computational costs while maintaining generation capabilities. ControlNet~\cite{zhang2023adding} and T2I-Adapter~\cite{ye2023ipadapter} introduced additional convolutional layers to incorporate conditional signals such as edges, poses, sketches, and segmentation, enabling controlled generation. These input conditions enhance \zp{task-specific generation by providing contextual guidance for image synthesis.} 
\cite{blattmann2023align, wu2023tune} have integrated the temporal dimension into diffusion models and finetuned them, expanding their application to video generation. \ly{In particular, several studies have expanded the text-guided control models into an image-controllable generative model using image conditioning~\cite{zhang2023magicavatar,ma2023follow}.
Despite their outstanding adaptation abilities at the image level, the aforementioned methods focus only on the controllability of the human subject within input images,} 
\zpn{limiting their output to basic 2D motions (\eg, frontal dancing).}

\noindent\textbf{Human Animation.}
\ly{Human} animation, which aims to generate images or videos from one or multiple input images, has become a crucial aspect of video generation. The integration of diffusion models has significantly advanced this field due to their superior generation quality and stable controllability. For example, PIDM~\cite{bhunia2023person} uses texture diffusion blocks to input desired texture patterns into the SD denoising process for human pose transfer. Similarly, DreamPose~\cite{karras2023dreampose} utilizes a pretrained stable diffusion model and proposes an adapter to model the CLIP~\cite{radford2021learning} image embeddings. 
DisCo~\cite{disco}, inspired by ControlNet, innovatively decouples the control of pose and background, providing finer control over the animation process, while also introducing artifacts and jittering in the characters' edges, since it did not reasonably integrate the foreground and background.
Animatediff~\cite{guo2023animatediff, guo2023sparsectrl} improves motion continuity by incorporating temporal layers, addressing jitter-related issues between frames. Champ~\cite{champ} attempts to use 3D representation SMPL and rarely learns the multi-view relationship between SMPL and human appearance. Despite these advancements, challenges such as texture inconsistency and temporal instability persist. Additionally, there is a need for methods that achieve control over characters in 3D sequences and  \ly{configurable} scenes, demonstrating a more generalized capability in character animation.

\noindent\textbf{Free-view Video Generation.}
Significant advancements in 3D neural representations, including NeRF \cite{mildenhall2021nerf} and 3D Gaussian splatting \cite{kerbl20233d}, have inspired a range of research efforts \cite{hu2024gaussianavatar, li2024animatable, liu2021neural, peng2021animatable, weng2022humannerf} that model dynamic humans as pose-conditioned NeRFs or Gaussians, allowing highly detailed animatable 3D avatars. However, these methodologies often rely on fitting neural fields to either multi-view recordings or monocular videos of dynamic subjects, which imposes severe limitations on their usability due to inefficient training processes and the significant resources needed for data acquisition. Recently, various studies \cite{anymateanyone, disco, magicanimate, champ} have examined the potential of 2D diffusion models but they are restricted to basic 2D movements within limited viewpoints. Human4Dit~\cite{shao2024human4dit360degreehumanvideo} attempts to synthesize 3D motions from a free-view perspective. It employs Transformers to establish the connection between the camera and multi-view appearance, yet this approach imposes significant demands in terms of both dataset and computational consumption. 
\zpn{Therefore, developing a geometry-guided approach is essential for efficiently achieving free-view video generation.}

\section{Method}
\label{sec:Method}

\subsection{Latent Diffusion Models}

Our method is based on the Latent Diffusion Model (LDM)~\cite{rombach2021highresolution}, which applies the diffusion processes within a latent space. Initially, LDM requires training a VAE consisting of an encoder $E$ and a decoder $D$.
The diffusion process involves a variance-preserving Markov process that incrementally introduces noise to an initial latent representation $z_0$ over $T$ time steps, generating diverse noisy latent representations. The process can be expressed as follows:
\begin{equation}
    \label{eq:diffusion_forward}
    \boldsymbol{z}_t = \sqrt{\Bar{\alpha}_t}\boldsymbol{z}_0 + \sqrt{1- \Bar{\alpha}_t}\boldsymbol{\epsilon}, \quad \epsilon \sim \mathcal{N}(\boldsymbol{0},\boldsymbol{I}),
\end{equation}
where $\overline{\alpha}_t$, $t = {1, ..., T}$, represents the noise intensity at each time step.
Following the final iteration of the diffusion process, the condition distribution $q(\boldsymbol{z}_T \mid \boldsymbol{z}_0)$ closely approximates a standard Gaussian distribution denoted by $\mathcal{N}(\boldsymbol{0},\boldsymbol{I})$.

During the denoising phase, the model predicts the noise $\boldsymbol{\epsilon}_{\theta}(\boldsymbol{z}_t,t,\boldsymbol{c})$ at each time step, working backwards from $z_t$ to $z_{t-1}$. 
$\boldsymbol{\epsilon}_{\theta}$ denotes the neural network to predict the noise.
The training function is commonly employed as the Mean Squared Error (MSE) loss:
\begin{equation}\label{eq:mse}
    \mathbb{E}_{\mathcal{E}(I), c_{\text{text}},\epsilon\sim\mathcal{N}(0,1),t}\left[\omega(t) \lVert \epsilon-\epsilon_{\theta}(z_t, t, c_{\text{text}}) \rVert_{2}^{2} \right]
\end{equation}
where $c_{text}$ refers to the text embedding obtained from the CLIP. After the training process, the model is adapted by methodically reversing the noise, starting from a noisy state $z_T$ drawn from a Gaussian distribution $\mathcal{N}(\boldsymbol{0},\boldsymbol{I})$ and moving towards the original state $z_0$. 

\subsection{Textured SMPL Representation}
Expressing the motion that occurs in 3D space using a single reference image is challenging, particularly when it involves significant movements with pronounced deformations and a novel perspective appearance. \ly{To address this, we propose a new 3D representation with structured prior that \zpn{geometrically} ensures more accurate expression of complex 3D movements under free-view perspectives.}

\begin{figure}[!t] 
    \includegraphics[width=\linewidth]{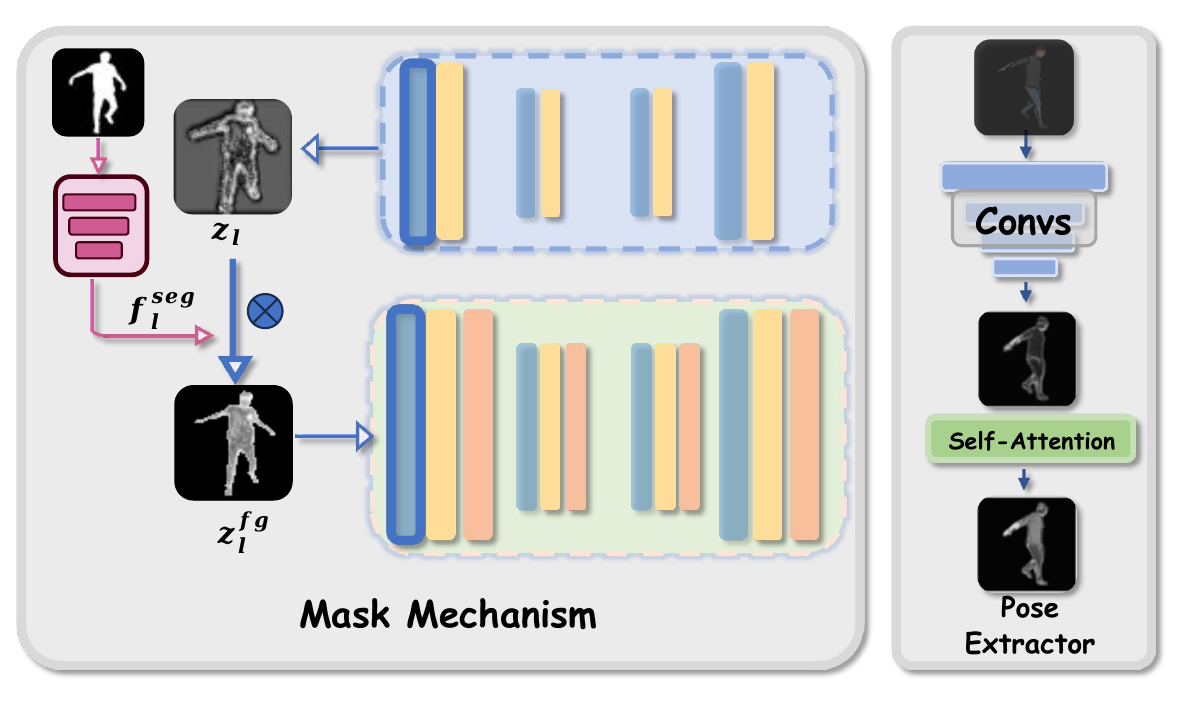}
     \vspace{-0.3in}
    \caption{\textbf{Implementation of the Masking Mechanism and Pose Extractor.} We visualize the operation of the masking mechanism and observe that features in the foreground region tend to diffuse toward the edges and overflow after the first layer of spatial attention. To mitigate this issue, we refine the foreground features using the downsampled $f_l^{seg}$. In the pose extractor, self-attention effectively captures \ly{structured information in SMPL representation}, including facial features, torso details, and clothing textures.
    }
    \label{fig:details}
\end{figure}

\noindent\textbf{Motion Signals.}
Given a reference image $I_{ref}$ and a desired sequence of 3D motions $I^{1:N}$, we aim to construct an adequate pose representation as motion control signals, which provides multi-view prior~\cite{3DControl} and ensures novel view generation with minimal training cost.

\ly{To achieve this}, \zp{we first extract the 3D human parametric SMPL model from a video sequence through the existing network~\cite{4Dhumans} or generated by language model~\cite{jiang2024motiongpt}.}
Then we employ a well-established methodology~\cite{casas2023smplitex} to construct a UV texture map $U_{part}$ for a user-desired character $I_{ref}$. 
Additionally, a human silhouette $s$ is calculated to mask the pixel-to-surface correspondence $d$, mapping each pixel $p \in I_{ref} $ onto the surface coordinates of a table using the map $d \otimes s$:
\begin{equation}
    U_{part} = \Pi(I_{ref}, d \otimes s)
\end{equation}
We perform inpainting on $U_{part}$ utilizing a frozen Stable Diffusion to obtain the final pseudo-complete UV map $U_{com}$, and overlay the $U_{com}$ onto the SMPL sequence as $\theta^{1:N}$. The textured SMPL sequence $\theta^{1:N}$ is warped into 2D space as $M^{1:N}$ through a user-defined camera trajectory $T^{1:N}$ to control the generation of the stable diffusion network:
\begin{equation}
M^{i} = \Omega (U_{com}, \theta ^{i} \cdot T^{i})  
\end{equation}

\noindent\textbf{Pose Extrarctor.}\label{sec:Pose Extrarctor}
\ly{Unlike the previously abstract control signals like skeleton~\cite{anymateanyone, wang2024unianimate}, densepose~\cite{dreampose, magicanimate}, and etc., this more structured prior requires additional processing of pixel information.}
In contrast to introducing any additional ControlNet, we have implemented a method similar to the condition encoder in~\cite{zhang2022exploring}, extracting appearance consistency prior from the pixel level to the latent space. \zp{Specifically,} we propose a pose extractor $\mathcal{P}$ combined a four-layer convolution to unify the dimensions with noise and an attention layer for capturing RGB features:
\begin{equation}
 \boldsymbol{z}_{pose} = \mathcal{P}(M^{i})
\end{equation} 

After that, $\boldsymbol{z}_{pose}$ is concatenated with the latent noise and fed into a 3D convolution layer for fusion and alignment. This approach utilizes effective priors, eliminating the need for extensive \zp{well-captured} videos to learn complex occlusion and alignment processes. Consequently, it significantly enhances model training efficiency while maintaining the fidelity of free-viewpoint human motion.

\subsection{Foreground-background Separation Learning}

To synthesize more realistic videos, we incorporate scene components into controllable attributes. Previous works \cite{anymateanyone, magicanimate, wang2024unianimate, tu2024motionfollower, wang2024vividpose} fail to model background interactions, changes, and dynamic transitions. DisCo \cite{disco} simply repeats a single image as the input, resulting in unsmooth background movement. Furthermore, it employs a basic mask to blend the foreground and background, causing noticeable flickering at their edges. We argue that these challenges arise from the limitations of the video attribute parser, which operates solely in a fully 2D feature space and overlooks the inherent spatial properties of a frame. To address this issue, we decompose a frame into a human foreground and a scene background, encoding them separately in \zp{the} latent space.

\begin{figure*}[!t] 
    \centering
    \includegraphics[width=\textwidth]{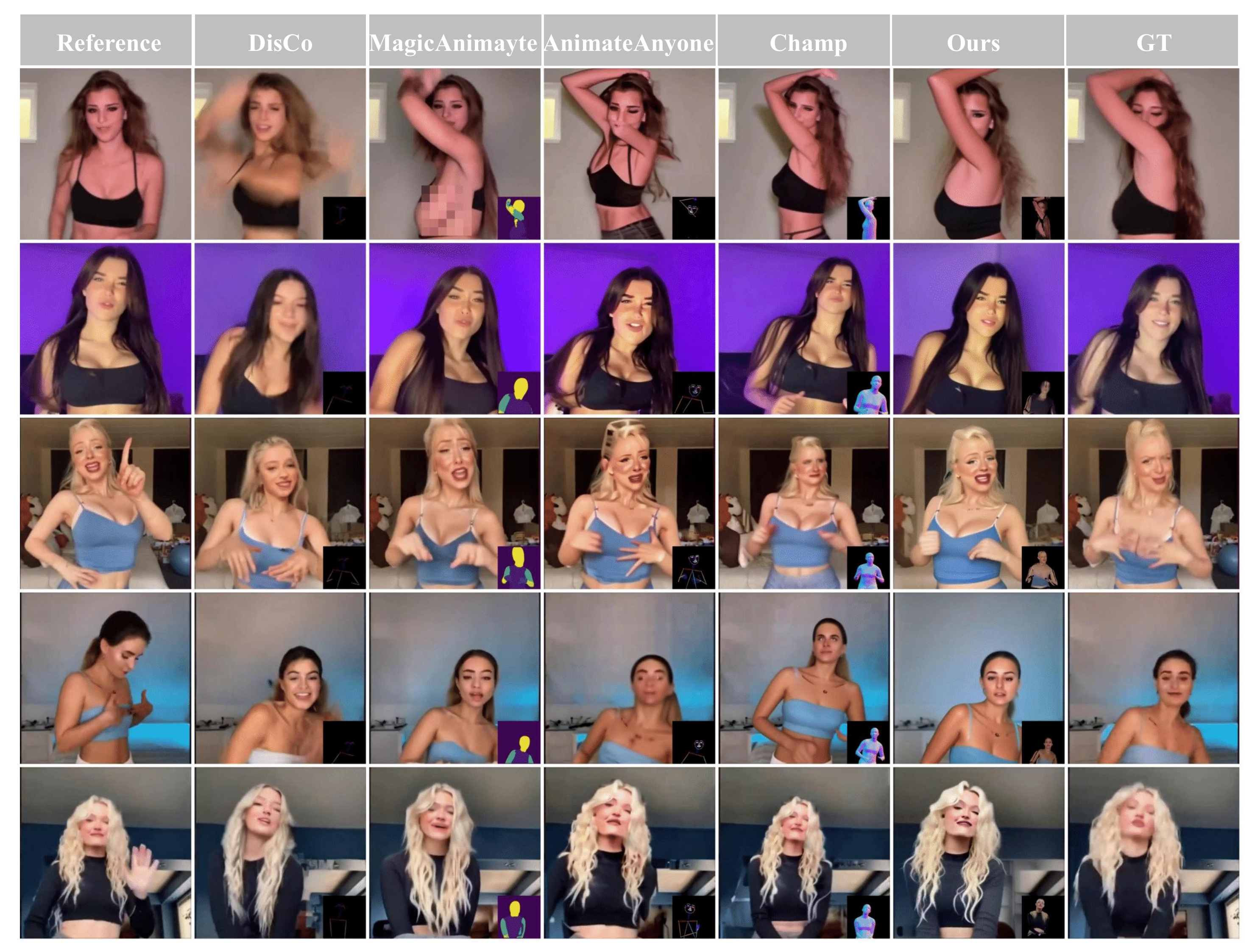}
     \vspace{-0.3in}
    \caption{
    \textbf{Qualitative comparisons between our approach and state-of-the-art methods on the TikTok dataset}. We annotate the control conditions in the bottom right corner. \ly{The SMPL representation provides robust priors that ensure the best reliability of appearance quality.}
    }
    \label{fig:tiktok_compare}
\end{figure*}

\noindent\textbf{Foreground Feature Extraction.}
We use the reference image $I_{fg}$, with the background removed, as the input to extract appearance features through spatial attention layers $l = 1,2,..., L$.
These features are processed by a model that shares the same architecture as the denoising U-Net, with \ly{all other parameters kept frozen}. The weights are initialized from a pre-trained Stable Diffusion model~\cite{musepose}.
We operate at different spatial resolutions $(h_l, w_l)$ to obtain foreground latents $\boldsymbol{z}_l \in \mathbb{R}^{(h_l\times w_l)\times c_l}$. The first half of the $\boldsymbol{z}_l$ is injected to the denoising U-Net by concatenating it with the noise latent along the spatial dimension~\cite{anymateanyone}.  This process aims to learn the correspondence between multi-view appearances and the color prior based on the SMPL representation $\boldsymbol{z}_{pose}$, thereby guiding spatial attention to encode the appropriate latent codes $\boldsymbol{z}_l$.

\noindent\textbf{Masking Mechanism.}
Although the non-foreground areas contain no features before input, we observe that as the network deepens, features progressively diffuse outward toward the contour edges. This diffusion causes the foreground latent features to expand beyond their boundaries, encroaching on regions designated for background synthesis, as illustrated in Fig.~\ref{fig:details}. This contour conflict leads to flickering.

Inspired by \cite{sdinpainting}, we propose a masking mechanism to prevent undesired feature diffusion. We create a binarized mask of the reference image $I_{ref}$ and downsample it to the resolution corresponding to each spatial layer $l$ before injecting the information into the denoising U-Net. We define the foreground region as $f_l^{seg} \in \mathbb{R}^{(h_l\times w_l)\times c_l}$ and use it to filter out information that exceeds the boundaries, thereby obtaining precise foreground features as latent codes:
\begin{equation}
   \boldsymbol{z}_l^{fg} = \boldsymbol{z}_l \otimes  f_l^{seg}
\end{equation}
This approach guarantees that only pure foreground features $\{ \boldsymbol{z}_1^{fg}, \boldsymbol{z}_2^{fg}, ...,  \boldsymbol{z}_L^{fg} \} \in \mathbb{Z}_{fg}$ are utilized to fill the target appearance as guided by the pose condition. The ablation study in Sec.~\ref{sec:alb} provides clear examples illustrating the differences between employing and not employing the masking strategy.

\noindent\textbf{Background Feature Extraction.} 
We integrate scenario sequences using a 'background encoder', that aligns with the structure of the SD encoder, specifically designed for encoding dynamic background sequences. These sequences are generated by removing human and occlusion objects from source videos or by rendering scene sequences in any 3D environment using camera trajectories $T^{1:N}$, as described in Sec.~\ref{sec:Pose Extrarctor}. We utilize a frozen VAE to map pixel information into the latent space, followed by a learnable background encoder to obtain the background latent representation $\mathbb{Z}_{bg}$. This sequence is then injected into the denoising U-Net by combining it with the noise latent during the decoding process.

\begin{figure*}[!t] 
    \centering
    \includegraphics[width=\textwidth]{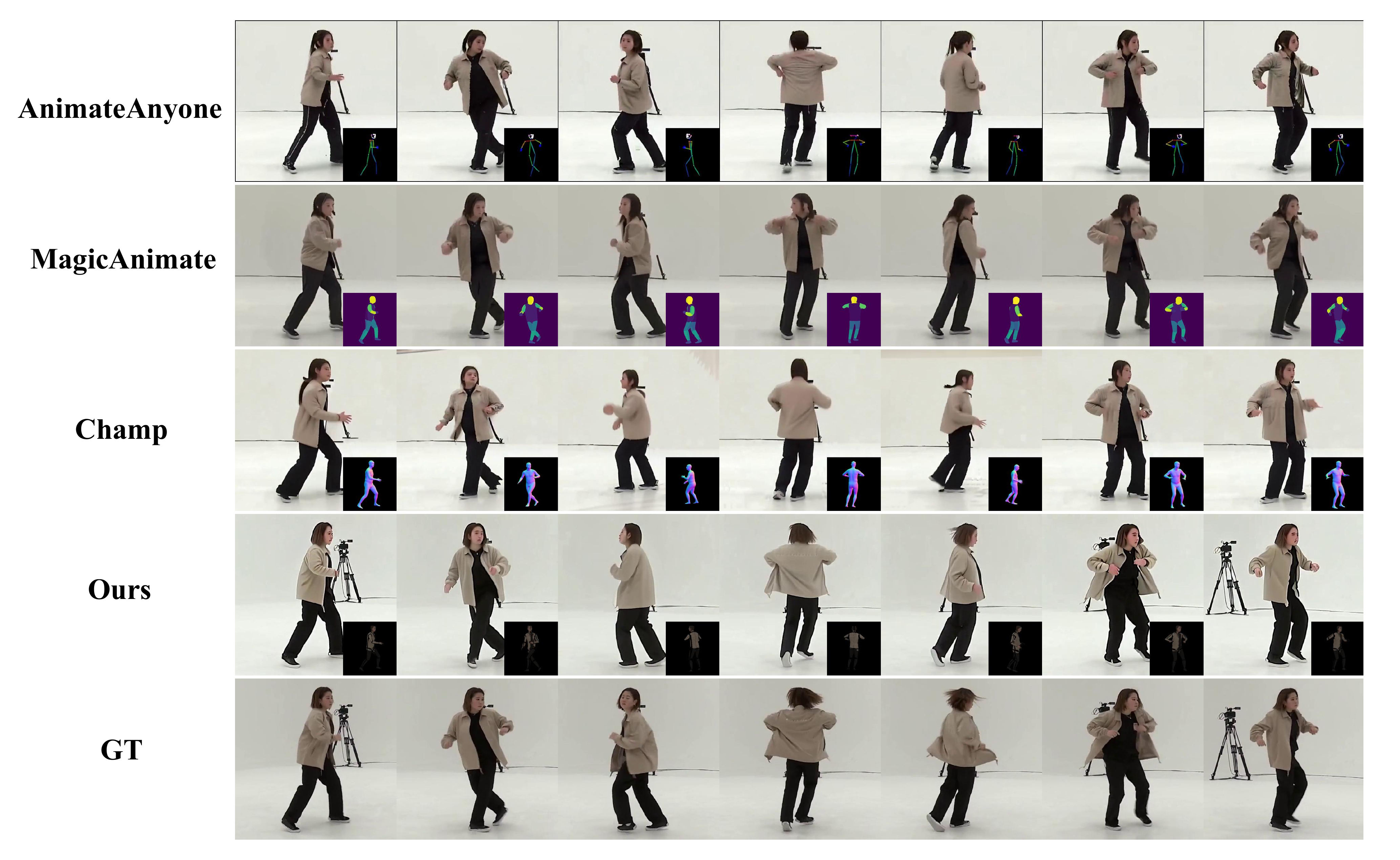}
    \vspace{-0.3in}
    \caption{
    \textbf{Qualitative comparison with state-of-the-art methods on the AIST dataset.} Our approach demonstrates the best quality in preserving both the fidelity and consistency of character appearance across 360-degree views.
    }
    \label{fig:asit_compare}
    \vspace{-0.05in}
\end{figure*}

\noindent\textbf{Composed Decoding to Recompose Latent Codes.}
Given the latent codes of decomposed attributes, we recompose them as conditions for the diffusion-based decoder in video generation. As illustrated in Fig.~\ref{fig:pipeline}, we employ a denoising U-Net backbone built on a pre-trained Stable Diffusion model, incorporating temporal layers from \cite{guo2023animatediff}. The identity condition for cross-attention is embedded through CLIP\cite{radford2021learning}, where we utilize the foreground image $I_{fg}$ to extract human embeddings $\mathbb{F}_{clip}$. When obtaining the fused noise latent $\mathbb{Z}_{encoder}$ concatenated with $\mathbb{Z}_{fg}$ after encoder processing, we freeze all U-Net layers except for the cross-attention. The update rule can be expressed as follows:
\begin{equation}
    \mathbb{Z}_{full} = \lambda\text{Softmax}\left(\frac{\textbf{Q}\textbf{K}_{bg}^T}{\sqrt{d}}\right)\textbf{V}_{fg} + \text{Softmax}\left(\frac{\textbf{Q}\textbf{K}_{noise}^T}{\sqrt{d}}\right)\textbf{V}_{fg}, 
\end{equation}
where $\textbf{Q} = \mathbb{Z}_{encoder}\textbf{W}_{\textbf{Q}}$, $\textbf{K}_{bg} = \mathbb{Z}_{bg}\textbf{W}_{\textbf{K}}$, $\textbf{K}_{noise} = \mathbb{Z}_{encoder}\textbf{W}_{\textbf{K}}$, $\textbf{V}_{fg} = \mathbb{F}_{clip}\textbf{W}_{\textbf{V}}$, and $\lambda$ is a hyperparameter defined as 1 here. 
We utilize the restriction loss to guide the background encoder in producing suitable embeddings for recomposition with the final noise and to facilitate the learning of cross-attention parameters.

\subsection{Training.}
We configure the denoising U-Net 
models with the motion module, initializing the pretrained weights from Musepose~\cite{musepose}. During training, we only optimize the pose extractor, spatial attention layers in the foreground encoder, cross-attention layers in the denoising U-Net, and the background encoder, while keeping the rest of the network's weights fixed. The following loss function is employed:
\begin{equation}\label{eq:loss}
    \mathcal{L} = \mathbb{E}_{x_0, z_{\text{fg}}, z_{\text{bg}}, z_{\text{pose}}, \epsilon\sim\mathcal{N}(0,1), t}\left[\lVert \epsilon-\epsilon_\theta({x_t, z_{\text{fg}}, z_{\text{bg}}, z_{\text{pose}}, t}) \rVert_{2}^{2} \right], 
\end{equation}
where $x_0$ represents the augmented input sample, $t=1,...,T$, denotes the diffusion timestep, $x_t$ is the noised sample at timestep at $t$, and $\epsilon_\theta$ signifies the function of the denoising U-Net.

\begin{table}[!t]
\tabcolsep 4pt
\centering
  \caption{\textbf{Quantitative results for human dance generation}. L1 is measured in units of E-04. Despite the high cost of learning, our approach exhibits significant advantages even on the TikTok dataset.
  }
  \resizebox{1.0\linewidth}{!}{
  \begin{tabular}{lp{1cm}<{\centering}p{1cm}<{\centering}p{1cm}<{\centering}p{1cm}<{\centering}p{1cm}<{\centering}p{1cm}<{\centering}}
    \toprule
   \multicolumn{1}{c}{\multirow{2}{*}{Methods}} & \multicolumn{4}{c}{Image} & \multicolumn{2}{c}{Video} 
    \\ \cmidrule(lr){2-5} \cmidrule(lr){6-7}  
    \multicolumn{1}{c}{} & \multicolumn{1}{c}{L1$\downarrow$} & \multicolumn{1}{c}{PSNR$\uparrow$} &  \multicolumn{1}{c}{SSIM$\uparrow$} &  \multicolumn{1}{c}{LPIPS$\downarrow$} &  \multicolumn{1}{c}{FID-VID $\downarrow$} & \multicolumn{1}{c}{FVD $\downarrow$} \\ 
    \midrule
        DreamPose\cite{dreampose}  & 6.91 &13.19 & 0.509 & 0.450 & 79.46 &  551.56 \\
        Disco\cite{disco} & 3.78 & 29.03 & 0.668 & 0.292 & 59.90 &  292.80 \\
        AnimateAnyone\cite{anymateanyone} & —— &  29.56 & 0.718 & 0.285 &  —— &  171.90  \\
        MagicAnimate\cite{magicanimate} & 3.13 & 29.16 & 0.714 & 0.239 & 21.75 & 179.07  \\
        Champ\cite{champ} & 2.94 &  29.84 & 0.773 & 0.235 & 26.14 & 170.20 \\
        MotionFollower\cite{tu2024motionfollower} & 2.89 & 29.25 & 0.793 & 0.230 & 22.36 & 159.88  \\
        UniAnimate\cite{wang2024unianimate} & 2.66 &  \textbf{30.77} & 0.811 & 0.231 &  —— &  148.06  \\
        VividPose\cite{wang2024vividpose} & 6.79 &  30.07 & 0.774 & 0.258 & 15.12 & 139.54 \\
        Ours &  \textbf{0.54} & 30.40 & \textbf{0.820} & \textbf{0.200} & \textbf{14.71} &  \textbf{130.46}  \\
  \bottomrule
  \end{tabular}
  }
  \label{tab:tiktokres}
  \vspace{-0.1in}
\end{table}

\section{Experiment}

\subsection{Implementation Details}

\begin{figure*}[!t] 
    \centering
    \includegraphics[width=\textwidth]{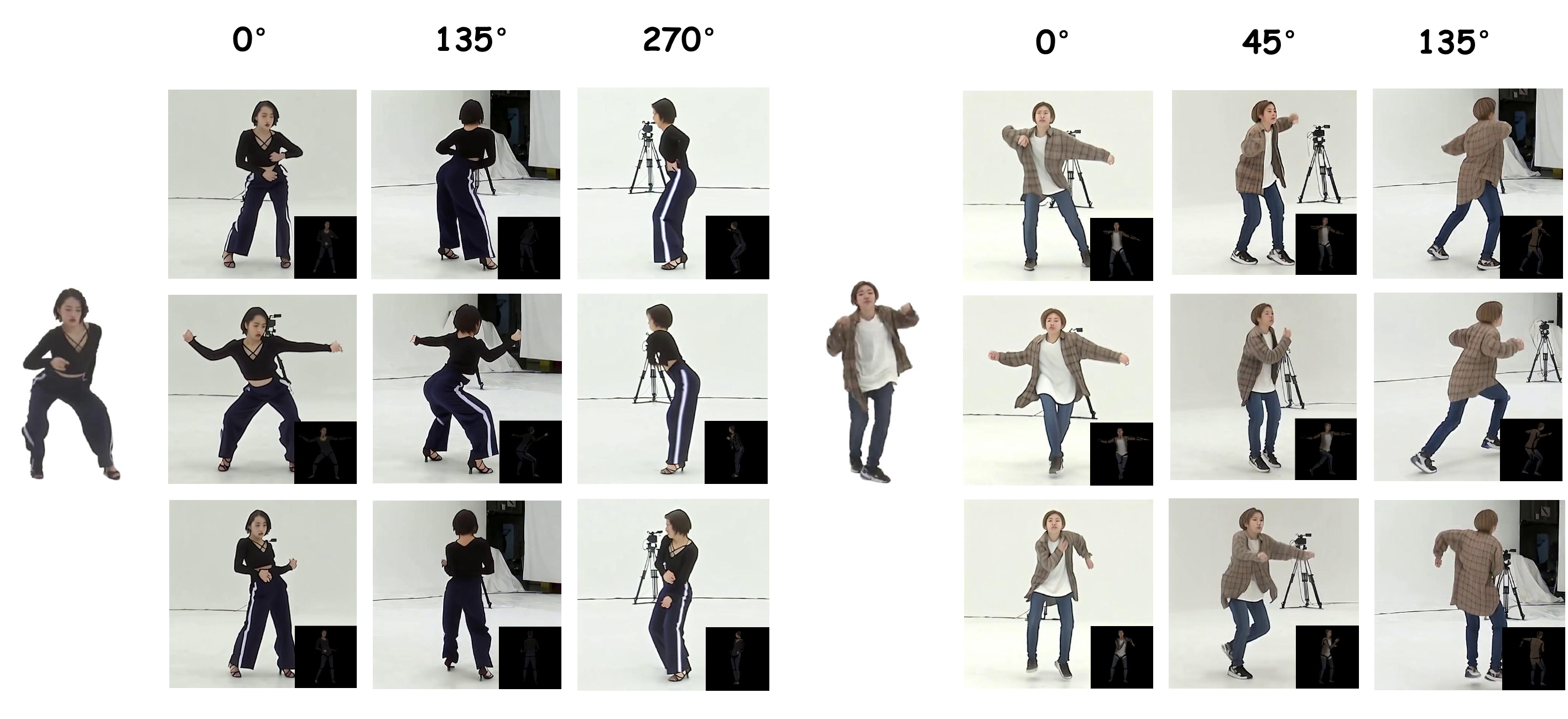}
    \vspace{-0.25in}
    \caption{
        \textbf{Qualitative results on multi-view videos.} Our method generates consistent multi-view videos from a single image without exhibiting appearance artifacts.
    }
    \label{fig:multi-view}
\end{figure*}

\noindent\textbf{Dataset.} 
We trained the core components using a small yet high-quality dataset that includes AIST and TikTok, featuring a variety of dance movements, scenes, and perspectives to ensure data diversity. The AIST Dance dataset~\cite{aist-dance-db} provides a rich source of multi-view information, capturing motion sequences from diverse individuals across nine distinct camera angles. We selected a diverse range of dance videos for each character, totaling 270 videos. Additionally, we curated a TikTok dataset~\cite{TikTok}, organized according to DisCo's training strategy. To enable accurate tracking of individuals within the video sequences, we employed GroundingDINO~\cite{groundingdino}, which simulated camera movement. For effective foreground-background separation, we utilized SGHM~\cite{SemanticGuidedHumanMatting}.

\noindent\textbf{Details.} 
First, we utilized SMPLitex~\cite{casas2023smplitex} to extract textures of human subjects from reference image and accurately warp them onto the SMPL model, aligned with 4D-humans\cite{4Dhumans}. In the initial training phase, character video frames underwent a series of preprocessing steps \ly{(\ie, resizing, sampling and center-cropping) resulting in a final resolution of 512$\times$512}. The training was conducted on 4 NVIDIA V100 GPUs, requiring approximately 30,000 iterations with 24 video frames and a batch size of 6 to achieve convergence. The learning rate was set to 1e-5. We trained the temporal layer for 10,000 iterations with a batch size of 1. During inference, we implemented the temporal aggregation strategy proposed in \cite{magicanimate}. To ensure fair and accurate comparisons, both the TikTok and AIST datasets were used as benchmarks.

\noindent\textbf{Metrics.} 
We rigorously evaluate the quality of the generated frames using established metrics from prior research. Following the methodology employed in DisCO~\cite{disco}, we report the average values of several key evaluation metrics, including PSNR, SSIM~\cite{wang2004image}, FID~\cite{unterthiner2018towards}, LPIPS~\cite{zhang2018perceptual}, and L1 error~\cite{janocha2017loss}. While these metrics primarily assess individual frames, we also incorporate FVD~\cite{unterthiner2018towards} and FID-VID~\cite{disco} to evaluate the perceptual quality and consistency of video sequences, thereby providing a comprehensive overview of our qualitative results.

\subsection{Qualitative Results}

\noindent\textbf{2D Human Motion.} 
We conducted experiments on the widely recognized monocular motion dataset TikTok~\cite{TikTok}. Fig.~\ref{fig:tiktok_compare} and Table~\ref{tab:tiktokres}  illustrate both the qualitative and quantitative results. 
Networks based on 2D control conditions, such as skeletons, can suffer from distortion or clipping issues in the foreground. Champ~\cite{champ}, which utilizes four different pose signals highly dependent on the normal map projected from the 3D model, is particularly sensitive to motion conditions, leading to artifacts and stiff motion distortions.
In contrast, Fig.~\ref{fig:tiktok_compare} demonstrates that, based on SMPL representation, yields character appearances with enhanced robustness and realism. 
\ly{While the separate input of foreground and background poses greater learning challenges, the edges remain exceptionally natural, resulting in an overall higher quality at the frame level.}
We have achieved exceptional results in both image generation and video metrics, as illustrated in Table~\ref{tab:tiktokres}. We cite results directly from \cite{champ} for DisCo, MagicAnimate, AnimateAnyone, and Champ, and DreamPose results from\cite{disco}.

\begin{table}[!t]
\tabcolsep 4pt
\centering
  \caption{\textbf{Comparison on the 3D human motion AIST dataset.} L1 is measured in units of E-04. 
  }
  \resizebox{1.0\linewidth}{!}{
  \begin{tabular}{lp{1cm}<{\centering}p{1cm}<{\centering}p{1cm}<{\centering}p{1cm}<{\centering}p{1cm}<{\centering}p{1cm}<{\centering}}
    \toprule
    \multicolumn{1}{c}{\multirow{2}{*}{Methods}} & \multicolumn{4}{c}{Image} & \multicolumn{1}{c}{Video} 
    \\ \cmidrule(lr){2-5} \cmidrule(lr){6-6}  
    \multicolumn{1}{c}{} & \multicolumn{1}{c}{L1$\downarrow$} & \multicolumn{1}{c}{PSNR$\uparrow$} &  \multicolumn{1}{c}{SSIM$\uparrow$} &  \multicolumn{1}{c}{LPIPS$\downarrow$} & \multicolumn{1}{c}{FID$\downarrow$}\\ 
    \midrule
         AnimateAnyone & 1.11 & 28.80  & 0.788 & 0.260 & 30.83  \\
         MagicAnimate & 1.67 &  28.13 & 0.751 & 0.314 &  62.94 \\
         champ & 1.63 & 28.94  & 0.726 & 0.339  & 43.74  \\
         Ours &  \textbf{0.63}  & \textbf{29.23} & \textbf{0.887} & \textbf{0.128} & \textbf{28.10}  \\
  \bottomrule
  \end{tabular}}
  \label{tab:asitres}
\end{table}

\noindent\textbf{3D Human Animation with Free Perspective Viewpoint.}
We divided the AIST dataset into 180 videos for training and 90 videos for testing. Each video is sampled at a frame rate of 20 FPS, resulting in a total of 100 frames per video. The first frame of the frontal view is used as the reference image. For Disco, MagicAnimate, Champ, and AnimateAnyone, we treat each view's video as a monocular video and perform inference separately after fine-tuning the motion module. \ly{Table~\ref{tab:asitres} presents a comprehensive overview of the results. As shown, our network achieves state-of-the-art performance in handling full-body, high-intensity movements.}

Our approach demonstrates remarkable superiority in tasks involving 3D animated full-body motion and multi-viewpoint conversion, thanks to the informative priors provided by our newly designed SMPL representation. Additionally, it effectively integrates the dynamic background with the foreground, producing more realistic and coherent generated videos. For specific comparison results, please refer to Fig.~\ref{fig:asit_compare}.

\begin{table}[!t]
    \centering
     \caption{\textbf{Comparison of SMPL Representation and Dwpose.}}
    \begin{tabular}{cccc}
         \toprule
           &  FID-VID     & FVD  & L1 \\
    \midrule
    dwpose & 13.17  &   311.96  & 3.73E-04    \\
    ours    & \textbf{11.99}  & \textbf{238.55}   & \textbf{6.33E-05} \\
    \bottomrule
    \end{tabular}
    \label{tab:smpl_representation_alb}
\end{table}

\begin{figure}[!t] 
    \centering
    \includegraphics[width=\linewidth]{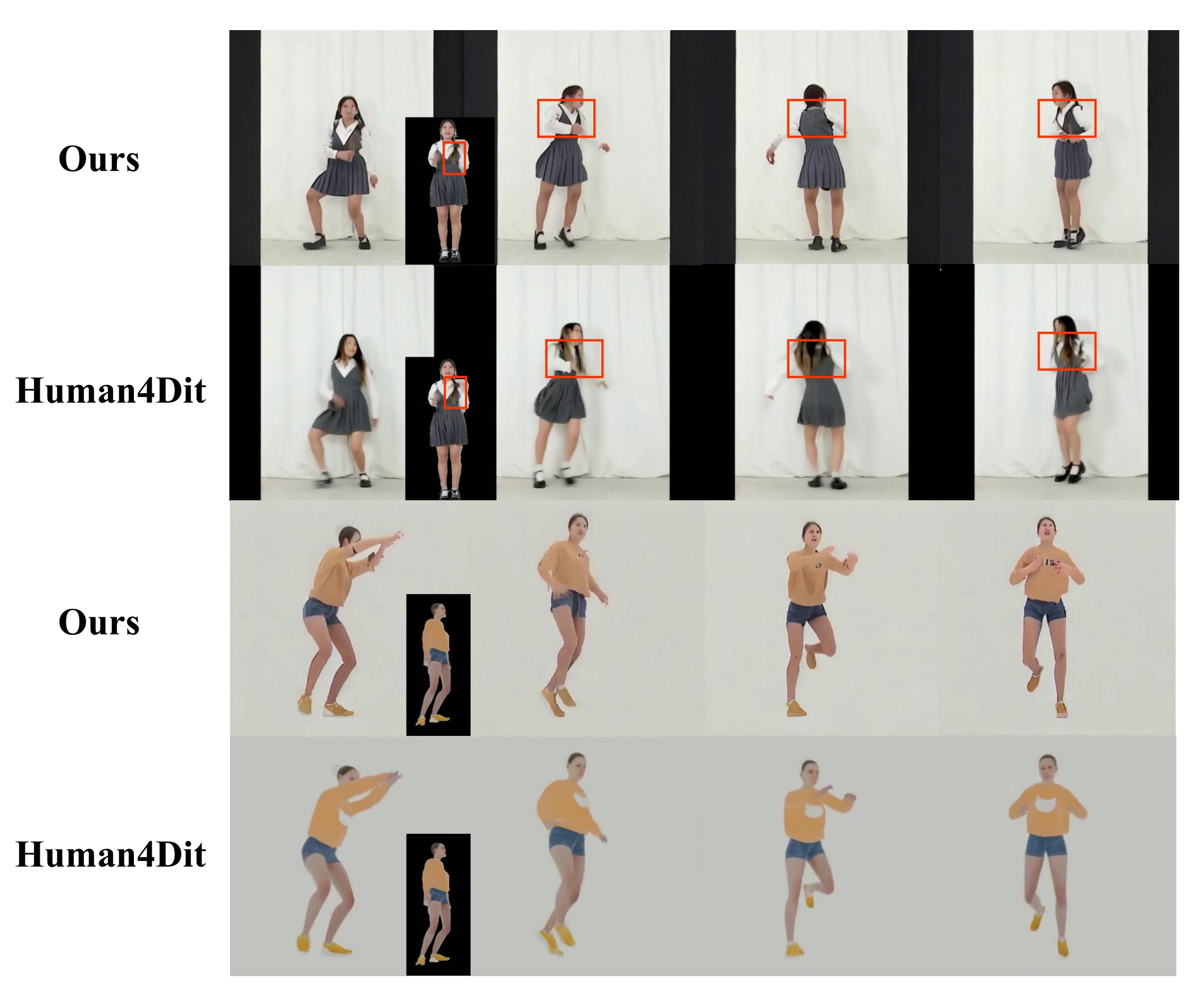}
     \vspace{-0.25in}
    \caption{
     \textbf{Comparisons on 360-degree video with Human4Dit.} \ly{Our method maintains higher fidelity to the reference image's appearance and character details across all viewpoints.}
    }
    \label{fig:human4dit}
\end{figure}

\begin{figure*}[!htbp] 
    \includegraphics[width=\linewidth]{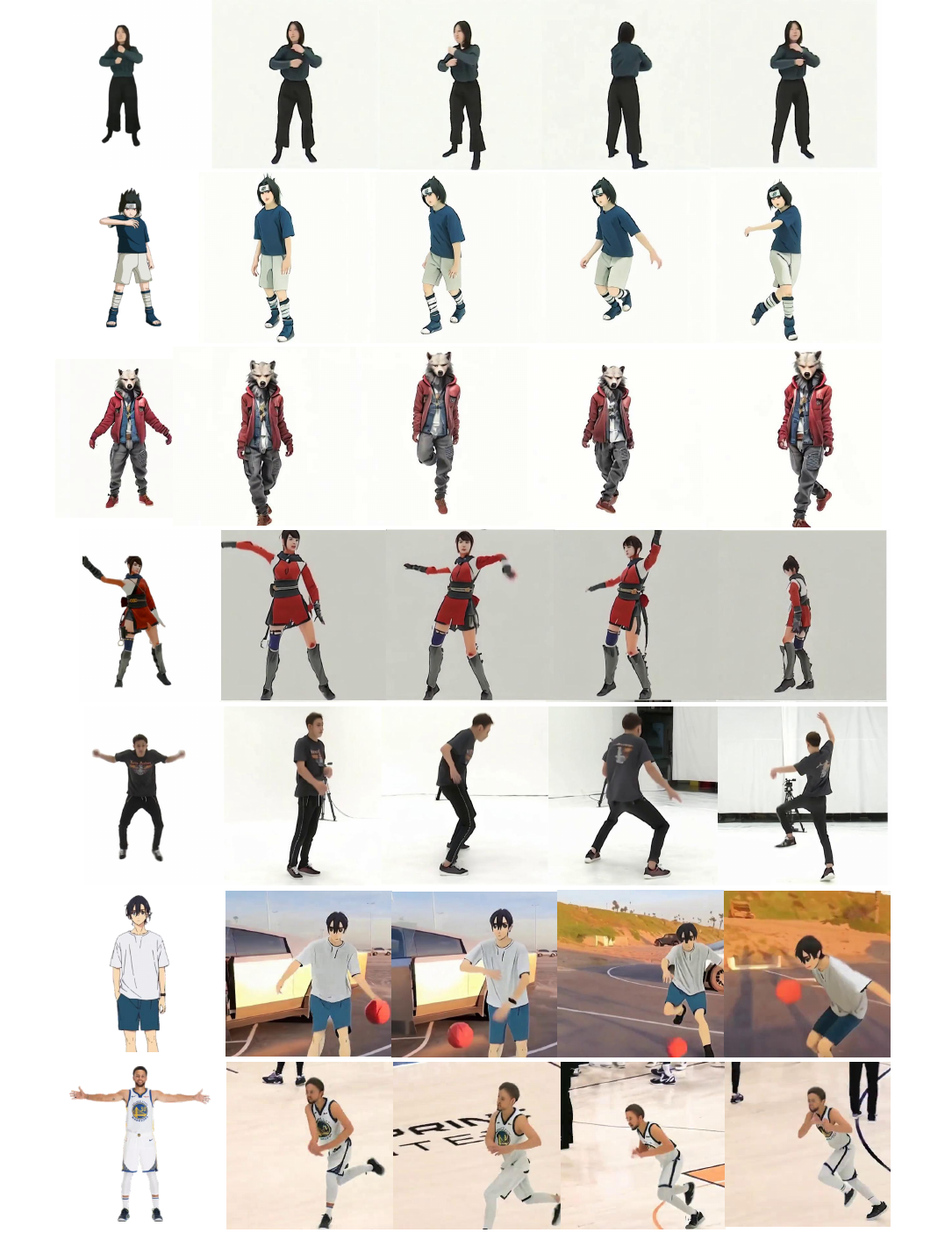}
    \caption{
    \ly{\textbf{Results on 4D in-the-wild data.} \method\ not only achieves human animation but also offers advanced capabilities for human video synthesis, including free-view motion transfer and user-desired scene insertion.} Some facial details are missing due to insufficient face information in reference image. Please refer to the facial generation capability shown in Fig.~\ref{fig:tiktok_compare}
    }
    \label{fig:4d_res}
\end{figure*}

To provide a more comprehensive comparison, we evaluated our method, \method, against the latest free-view approach Human4Dit\cite{shao2024human4dit360degreehumanvideo} (non-open-source) using free-view data in Fig.~\ref{fig:human4dit}. We directly reference results from the Human4Dit website, \ly{which serves as in-the-wild data for \method{}}. Leveraging structured prior representations, our method demonstrates improved fidelity to the appearance of images in novel views. For instance, in the case of the first young girl, we can maintain consistent hair styling from any angle while enhancing the details of her dress folds and overall character fidelity, whereas Human4Dit exhibits noticeable distortion. The girl's hair is flowing down loosely, instead of matching the style seen in the reference image.
In addition, we conducted qualitative results for scenarios that involve transforming camera viewpoints while maintaining the same motion, as illustrated in Fig.~\ref{fig:multi-view}. The discrepancies between different viewpoints pose substantial challenges for existing methods. In contrast, our approach ensures high-quality generation for each independent viewpoint while maintaining consistency across different perspectives.

\noindent\textbf{4D Human Animation In-the-wild Data.}
We generalize our model to in-the-wild data, synthesizing 4D videos from arbitrary motions and backgrounds following the same camera trajectory. Our method overcomes the limitations of 2D input and view constraints, successfully generating high-quality videos from any perspective. Despite training on a small dataset, we achieved effective correspondence between single images and their novel views, thanks to the stable 3D representation and efficient, reasonable texture prior. Moreover, the videos we produce carefully consider spatial coherence, incorporate realistic scene transitions, and effectively smooth out potential edge-related issues, refer to Fig.~\ref{fig:4d_res}.

\subsection{Ablation Studies}\label{sec:alb}
\begin{figure}[ht] 
    \makebox[\linewidth]{%
        \hfill \textbf{SMPL~~~~} \hfill \textbf{DWPose} \hfill \textbf{~~~~~~GT} \hfill
    }
    \includegraphics[width=\linewidth, trim=0 0 0 59, clip]{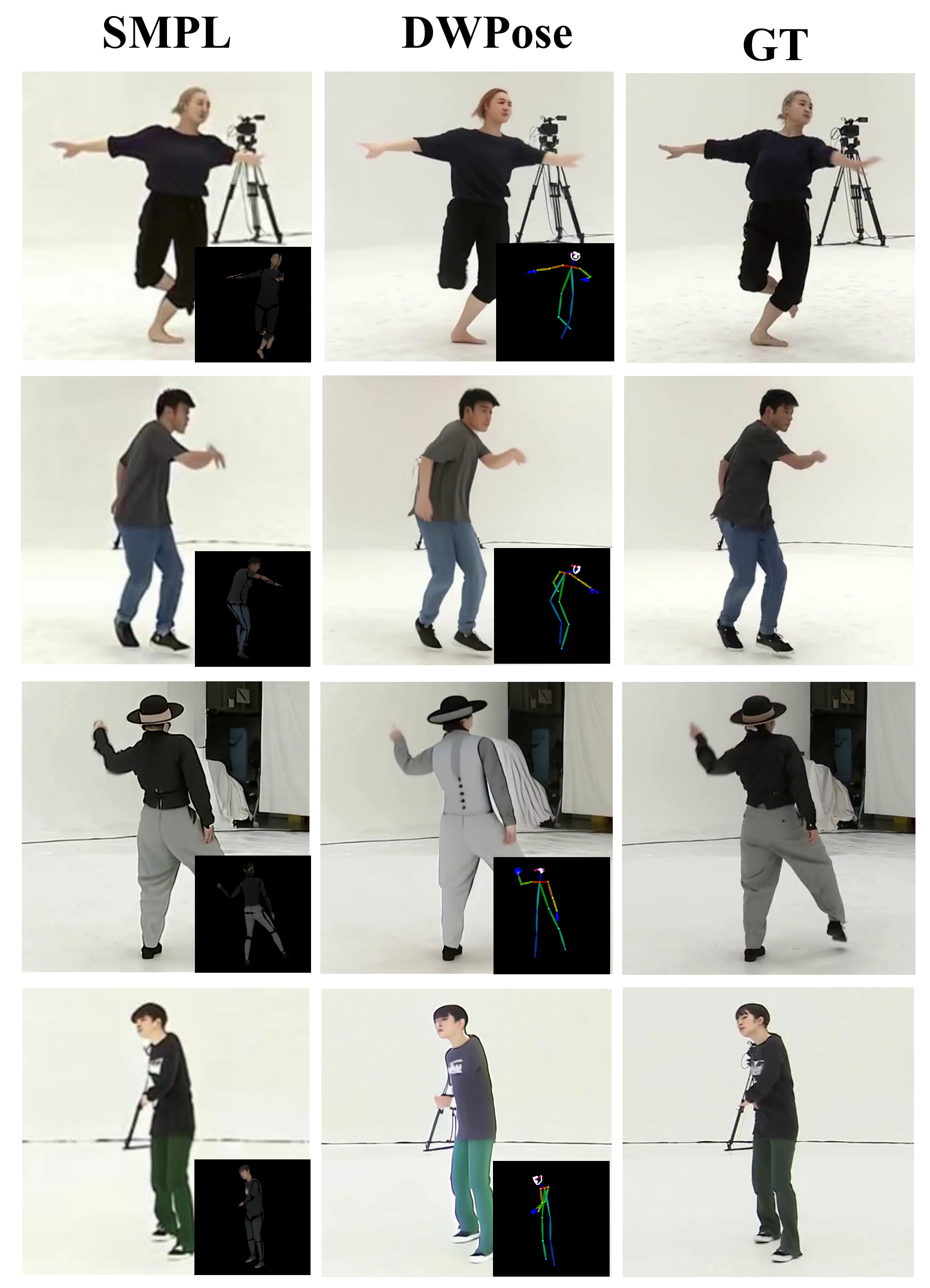}
    \caption{
    \textbf{Ablation results of the SMPL representation.} \ly{The SMPL representation significantly enhances both the geometric accuracy and appearance fidelity of characters, which can be attributed to its 3D representation and the use of texture priors.}
    }
    \label{fig:alb1}
\end{figure}

\noindent\textbf{SMPL Representation.}
To validate the effectiveness of the SMPL representation, we replaced the pose signals with the commonly used representation~\cite{dwpose}, to evaluate the impact of the texture prior. As shown in Fig.~\ref{fig:alb1}, using 2D pose conditioning in planar folding increases information entropy, compromising the accuracy of synthesized human motions. The limited identity information provided by the reference image restricts the ability to generate novel viewpoints in 3D motion, resulting in erratic oscillations during viewpoint transitions.

However, the use of SMPL representation produced more stable appearances, showing clear advantages in challenging multi-view scenarios. By leveraging texture priors, the synthesized frames not only preserve identity faithfully but also reasonably expand consistent novel views of humans with minimal cost. This greatly improves the network's versatility and applicability.

We compared the effectiveness of the SMPL representation using metrics such as FIV-VID, FVD, and L1, as shown in Table~\ref{tab:smpl_representation_alb}. The results demonstrate that incorporating texture priors significantly outperforms those utilizing DWPose. This enhancement is attributed to the SMPL representation's capacity to guide the learning process toward achieving more accurate appearances, both geometrically and at the pixel level.

\begin{figure}[ht] 
      \makebox[\linewidth]{%
        \hfill \textbf{w/o~~~~~~~} \hfill \textbf{w/~} \hfill \textbf{~~~~~~~GT} \hfill
    }
    \includegraphics[width=\linewidth, trim=0 0 0 50, clip]{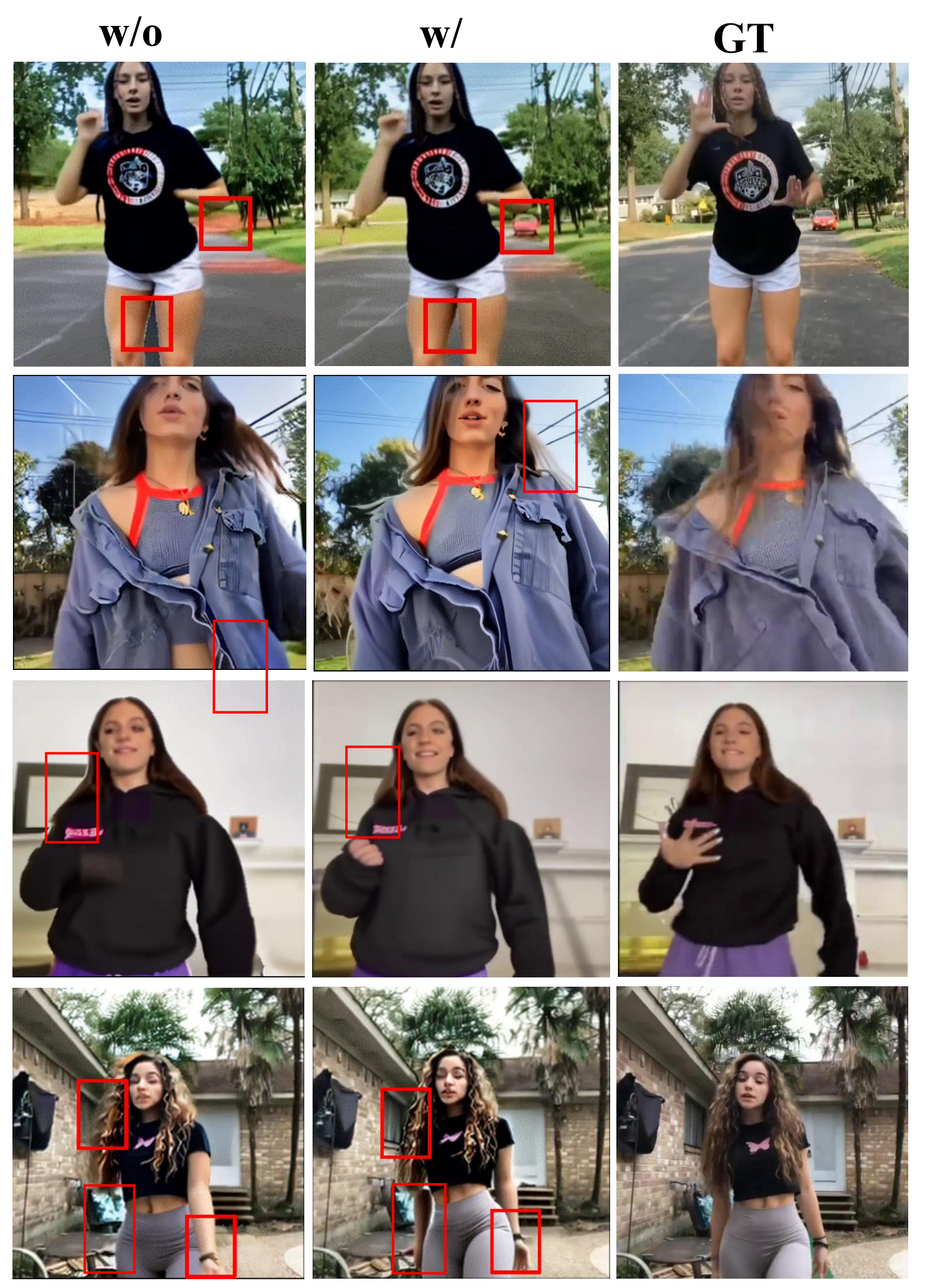}
    \caption{
    \textbf{Ablation results of the masking mechanism.} \ly{The masking mechanism effectively integrates the foreground and background, preventing flickering caused by conflicts between these regions.}
    }
    \label{fig:msking mchanism}
\end{figure}

\begin{table}[!t]
    \centering
    \caption{\textbf{Comparison with/without Masking Mechanism.}}
    \begin{tabular}{cccc}
        \toprule
          & PSNR     &  SSIM  & LPIPS     \\
    \midrule
    w/o  & 29.03  & 0.668  & 0.292       \\
   w/ &  \textbf{30.40}  & \textbf{0.820} & \textbf{0.200} \\
    \bottomrule
    \label{tab:abl2}
    \end{tabular}
\end{table}

\noindent\textbf{Masking Mechanism.}
\ly{We compared the results with and without the masking mechanism,  a critical technique for separate learning of foreground and background.} As shown in Fig.~\ref{fig:msking mchanism}, the addition of the masking mechanism significantly reduces edge jitter and blending artifacts, as highlighted in the red boxes. Specifically, the foreground area exhibits enhanced visual clarity and effectively minimizes unexpected jitter at the boundary between the human subject and the background. Furthermore, the contours provided by the foreground segmentation prior are noticeably smoothed, improving overall image quality. In line with these findings, the results in Table~\ref{tab:abl2} indicate that incorporating the masking mechanism results in higher-quality modified images.

\section{Contribution}
In this paper, we introduced \method, a novel framework \ly{for controllable human video synthesis that allows for flexible user control and addresses the limitations of previous animation networks in free-viewpoint manipulation and background \zpn{substitution}.} We implemented a texture-based SMPL representation, which provides color priors across various viewpoints, enhancing robustness in complex motion generation at minimal cost. Additionally, we developed a foreground-background separation learning strategy utilizing a masking mechanism, enabling hierarchical separation of spatial components throughout a video frame and facilitating proper recomposition. Experimental results demonstrate that our method not only allows for flexible control over characters, motions, and scenes but also offers advanced flexibility for arbitrary humans, generality to novel 3D motions, and applicability to interactive scenes.

\noindent\textbf{Limitations.} 
Unlike single-view approaches, we introduce simultaneous transformations of the foreground and background, which \zpn{may affect color interplay and introduce}  
chromatic aberrations in the generated images. Additionally, the estimation of the texture map solely relies on a single reference, \ly{leading to \zpn{possible} instabilities in generation quality across different perspectives.} 
\zpn{A potential improvement is to leverage the latest generative models to learn more robust texture priors and collect some realistic datasets, both with and without humans, to better study authentic relationships between foreground and background elements. }

\bibliographystyle{IEEEtran}
\bibliography{template}

\end{document}